%% file: _main.tex
\input{_constants}
\arxiv

\pdfoutput=1
\documentclass[10pt,twocolumn,letterpaper]{article}
\input{cvpr_header}

\unless\ifarxiv \myexternaldocument{_supplementary} \fi
\usepackage{pifont}
\newcommand*\colourcheck[1]{%
  \expandafter\newcommand\csname #1check\endcsname{\textcolor{#1}{\ding{52}}}%
}
\colourcheck{blue}
\colourcheck{green}
\colourcheck{red}
\usepackage{amsmath}

\newcommand{\xmark}{\ding{55}}

\begin{document}
\title{\paperTitle}
\author{\authorBlock}
\twocolumn[{%
\renewcommand\twocolumn[1][]{#1}%
\maketitle
\begin{center}
    \centering
       \includegraphics[width=.9\textwidth]{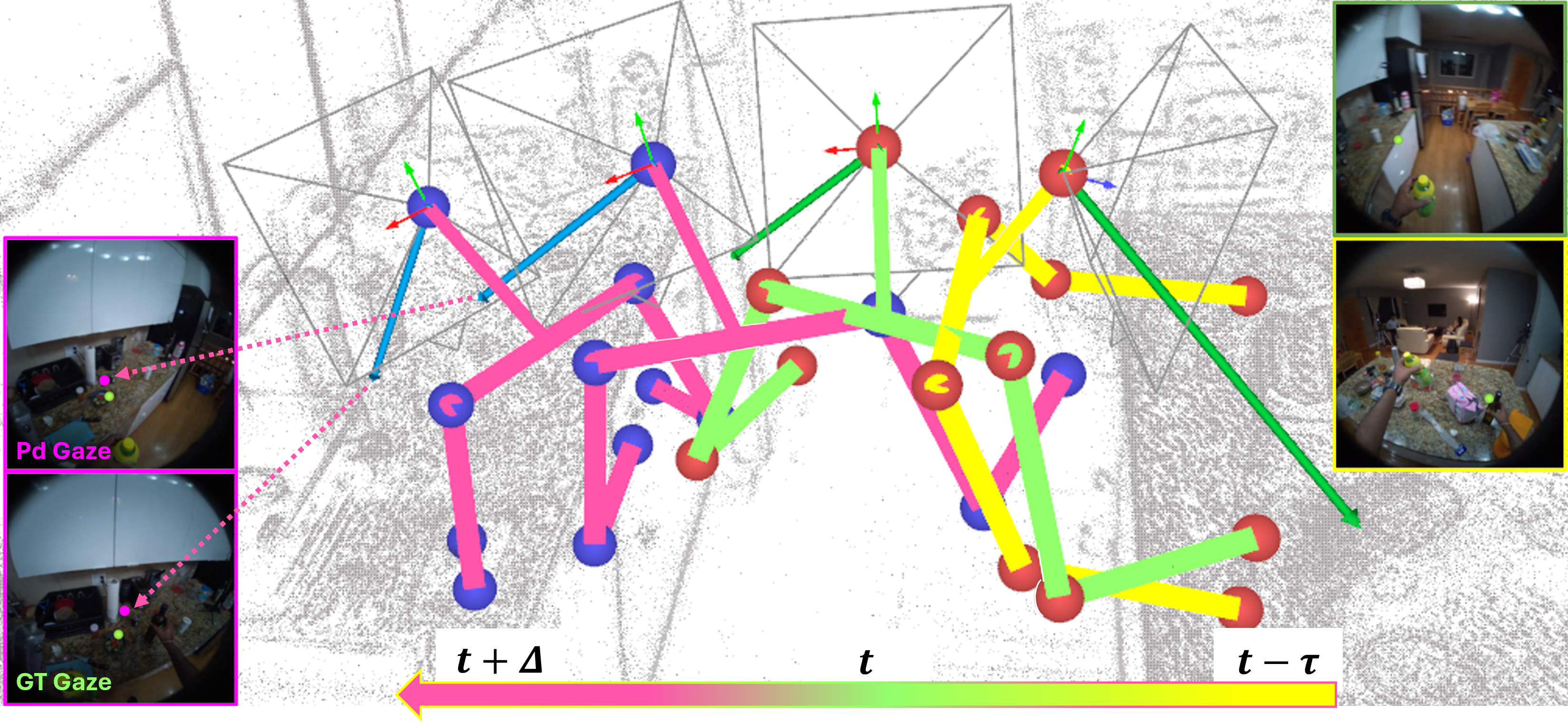}
        \captionof{figure}{We represent the Human Visuomotor System as a joint encoding of \textbf{\textit{Head Pose}}, \textbf{\textit{3D Gaze Direction}}, and \textbf{\textit{Upper-Body Joints}}. Given a sequence of visuomotor inputs and egocentric frames, our goal is to predict how the system coordinates its movements in the future. As forecasted egocentric frames are unavailable, predicted 2D gaze is mapped by finding the intersection of the gaze ray and the environment.}
    \label{fig:teaser}
\end{center}
}]

\maketitle

\input{00_abstract}
\input{01_intro}
\input{02_related}

\input{03_method}
\input{04_experiment}

\input{10_conclusion}

{\small
\bibliographystyle{ieeenat_fullname}
\bibliography{11_references}
}

\ifarxiv \clearpage \appendix

\input{12_appendix} \fi

\end{document}

%% file: _constants.tex
\def\paperID{1251}
\def\confName{ICCV}
\def\confYear{2025}

\def\paperTitle{Learning Predictive Visuomotor Coordination}

\def\authorBlock{
    Wenqi Jia$^{1}$, 
    Bolin Lai$^{2}$, 
    Miao Liu$^{3}$, 
    Danfei Xu$^{2\ast}$, 
    James M. Rehg$^{1\ast}$ \\
    $^1$University of Illinois Urbana-Champaign, $^2$Georgia Tech, $^3$Meta AI \\
    {\tt\small \{wenqij5,jrehg\}@illinois.edu, \{bolin.lai,danfei\}@gatech.edu, miaoliu@meta.com} \\
    {\small * Equal Advising}
    }

\newif\ifreview 
\newif\ifarxiv \newcommand{\arxiv}{\arxivtrue}
\newif\ifcamera 
\newif\ifrebuttal 

%% file: cvpr_header.tex
\ifreview \usepackage[review]{cvpr} \fi
\ifarxiv \usepackage[pagenumbers]{cvpr} \fi
\ifrebuttal \usepackage[rebuttal]{cvpr} \fi
\ifcamera \usepackage{cvpr} \fi

\input{_macros}  

\usepackage{xr-hyper}

\makeatletter
\newcommand*{\addFileDependency}[1]{
  \typeout{(#1)}
  \@addtofilelist{#1}
  \IfFileExists{#1}{}{\typeout{No file #1.}}
}

\makeatother
\newcommand*{\myexternaldocument}[1]{
    \externaldocument{#1}
    \addFileDependency{#1.tex}
    \addFileDependency{#1.aux}
}

\definecolor{cvprblue}{rgb}{0.21,0.49,0.74}
\usepackage[pagebackref,breaklinks,colorlinks,allcolors=cvprblue]{hyperref}
\usepackage[capitalize]{cleveref}
\crefname{section}{Sec.}{Secs.}
\crefname{table}{Table}{Tables}
\crefname{figure}{Fig.}{Figs.}

\ifarxiv \crefname{appendix}{App.}{Apps.}
\else \crefname{appendix}{Suppl.}{Suppls.} \fi

\usepackage{tikz}

%% file: _macros.tex
\usepackage{graphicx}	
\usepackage{amsmath}	
\usepackage{amssymb}	
\usepackage{booktabs}
\usepackage{times}
\usepackage{microtype}
\usepackage{epsfig}
\usepackage{caption}
\usepackage{float}
\usepackage{placeins}
\usepackage{color, colortbl}
\usepackage{stfloats}
\usepackage{enumitem}
\usepackage{tabularx}
\usepackage{xstring}
\usepackage{multirow}
\usepackage{xspace}
\usepackage{url}
\usepackage{subcaption}
\usepackage{xcolor}
\usepackage[hang,flushmargin]{footmisc}
\usepackage{svg}

\ifcamera \usepackage[accsupp]{axessibility} \fi

% ...

\ifarxiv  \fi

\newcommand{\R}[1]{{%
    \textbf{%
        \ifstrequal{#1}{1}{\textcolor{red}{R#1}}{%
        \ifstrequal{#1}{2}{\textcolor{blue}{R#1}}{%
        \ifstrequal{#1}{3}{\textcolor{magenta}{R#1}}{%
        \ifstrequal{#1}{4}{\textcolor{teal}{R#1}}{%
                           \textcolor{cyan}{R#1}%
        }}}}%
    }%
}}

%% file: 00_abstract.tex
\begin{abstract}
Understanding and predicting human visuomotor coordination is crucial for applications in robotics, human-computer interaction, and assistive technologies. This work introduces a forecasting-based task for visuomotor modeling, where the goal is to predict head pose, gaze, and upper-body motion from egocentric visual and kinematic observations. We propose a \textit{Visuomotor Coordination Representation} (VCR) that learns structured temporal dependencies across these multimodal signals. We extend a diffusion-based motion modeling framework that integrates egocentric vision and kinematic sequences, enabling temporally coherent and accurate visuomotor predictions. Our approach is evaluated on the large-scale EgoExo4D dataset, demonstrating strong generalization across diverse real-world activities. Our results highlight the importance of multimodal integration in understanding visuomotor coordination, contributing to research in visuomotor learning and human behavior modeling. Project Page: \href{https://vjwq.github.io/VCR/}{https://vjwq.github.io/VCR/}. 
\end{abstract}

%% file: 01_intro.tex
\section{Introduction}
\label{sec:intro}
Analyzing egocentric video to predict what the camera-wearer is going to do next is an important task in egocentric vision with applications in Augmented Reality (AR) and robotics. In order for an AI assistant in a pair of smart eyeglasses to be helpful, it should be able to anticipate what a person might be trying to do and provide advice or interventions \emph{before} a problem occurs. As a result, previous works have developed deep learning models to forecast ego-motion, gaze, and hand trajectories from egocentric videos \cite{jia2022generative,lai2022eye,li2023ego}. However, most prior methods focus on isolated modality signals (e.g., only gaze or hand motion) and fail to model their multimodal interdependencies. A key missing element in these prior works is a detailed consideration of the visumotor control system which underlies all goal-directed human movement. 

Psychological research underscores the predictive nature of visuomotor coordination, particularly in how visual memory informs motor planning. Hayhoe et al. \cite{hayhoe2003visual} observed that during natural tasks like making a sandwich, individuals rely on visual information from previous fixations to plan and coordinate movements over the course of several seconds. This reliance on stored visual memory allows humans to anticipate object locations and properties, ensuring seamless interaction with the environment. We hypothesize that understanding and modeling these predictive visuomotor behaviors can provide more effective forecaseting capabilities for robotics, VR/AR, and human-computer interaction, where anticipating human movement can enhance real-time decision-making and system adaptability.

Beyond egocentric vision, visuomotor prediction has also been explored in robotics to enhance manipulation and assistive tasks. Learning from human demonstrations \cite{finn2017one, lynch2019learning, zeng2020transporter} has enabled robots to generalize across diverse actions. However, these methods often rely on task-specific datasets and do not model full-body visuomotor coordination in natural settings. Bridging the gap between egocentric perception and full-body visuomotor learning remains an open challenge.  

We propose the first comprehensive predictive visuomotor learning framework, which jointly models head pose, gaze direction, and upper-body joint movements in 3D space. Unlike prior approaches that treat these signals independently, our model explicitly captures their temporal dependencies by encoding kinematic sequences and generating future motion trajectories using a diffusion model. By conditioning on multimodal observations, our method learns structured visuomotor patterns, enabling more temporally consistent and accurate movement predictions.

Our approach is enabled by the recent emergence of the EgoExo4D \cite{grauman2024ego} and Nymeria \cite{ma2024nymeria} datasets, which contain comprehensive 3D annotations and make it possible to quantitatively evaluate visumotor forecasting performance. Specifically, we evaluate our approach on EgoExo4D, leveraging its 3D annotations of gaze, head pose, and body motion. Averaged over a 1-second forecasting horizon, our model achieves an average error of 59 mm for visuomotor translation (head position, gaze ray endpoint, and upper-body joints) and a head rotation error of 13 degrees, demonstrating promising performance across a diverse set of daily activities. By jointly modeling these visuomotor signals, our method provides deeper insights into their temporal dynamics, advancing the study of predictive human motion modeling.
Our contributions are summarized as follows: 
\begin{itemize}
    \item We formulate a forecasting-based task for human visuomotor modeling, integrating head pose, gaze direction, fixation, and upper-body motion to capture temporal dependencies.
    \item We extend a diffusion-based framework that integrates egocentric vision and kinematic sequences, leveraging multimodal fusion to enhance temporal coherence and prediction accuracy.  
    \item We conduct extensive evaluations on the large-scale EgoExo4D dataset, demonstrating consistently strong performance across diverse real-world activities.  
    \item Our analysis includes extensive quantitative and qualitative evaluations, with a comprehensive ablation study assessing the impact of each multimodal component on accuracy.  
\end{itemize}

%% file: 02_related.tex
\section{Related Work}
\label{sec:related}

\subsection{Human Visuomotor Coordination}

Human visuomotor coordination is fundamental to action planning and execution, integrating first-person perspective visual perception, head orientation, and proprioception to enable fluid, goal-directed movements. Unlike purely reactive control, it operates \textbf{\textit{predictively}}, allowing for smooth and adaptive actions despite sensory and neural delays \cite{nijhawan2009compensating,wolpert2001motor,shadmehr2010error, zago2009visuo,avraham2019effects}.
Research in neuroscience and motor control highlights that head movement, gaze, and body posture work together to anticipate motion trajectories \cite{lappi2018visuomotor,land2001ways,bizzi1984posture}. Head orientation provides a stable spatial reference, aligning sensory input with motor execution, especially in dynamic environments\cite{graziano2006organization}. Proprioception further refines this process, allowing the brain to track limb positions and adjust movements accordingly\cite{shadmehr2008computational}. These multimodal cues are tightly coupled—humans naturally orient their head and upper body before executing reaching, stepping, or object manipulation tasks \cite{johansson2009coding}. This predictive integration is essential for both action preparation and real-time adaptation\cite{land2001ways,shadmehr2008computational}.

Inspired by these biological mechanisms, our work explicitly integrates head orientation, gaze, and egocentric perception into motion forecasting, not as independent signals but bridging neuroscience insights with real-world human movement prediction as a structured visuomotor coordination representation. 

\subsection{Modeling Predictive Visuomotor Coordination}

Existing human motion forecasting methods often focus on isolated sub-components rather than their integration. Gaze forecasting predicts future fixation points from past gaze trajectories, often using egocentric video \cite{zhang2017deep,lai2022eye,lai2023listen}, but it does not model how gaze directs motor actions. Hand trajectory forecasting predicts hand motion in first-person views, typically for object interactions \cite{liu2022joint,jia2022generative,grauman2022ego4d}, yet it largely ignores head movement and body posture, which naturally influence hand coordination. Full-body motion forecasting predicts skeletal motion based on past joint positions \cite{fragkiadaki2015recurrent,martinez2017human,pavllo2020modeling,mao2019learning,hernandez2019human,guo2023back,chen2023humanmac}, but lower-body motion is often dictated by external terrain constraints rather than internal visuomotor coordination. In contrast, upper-body motion is directly linked to fine motor tasks, object interactions, and skill-based activities, where visuomotor coordination plays a critical role. Thus, we focus on \textbf{\textit{upper-body dynamics}}, ensuring our approach remains independent of environmental priors while maintaining relevance across diverse skilled activities.  

Egocentric vision has emerged as a key modality for studying human visuomotor coordination, as it provides a first-person perspective of perception and action. Unlike third-person views, egocentric video directly captures the relationship between visual attention, head movement, and motor execution, making it well-suited for modeling predictive visuomotor behaviors \cite{zhang2017deep,jia2022generative,li2023ego,tan2023egodistill}. This perspective is particularly valuable for understanding how humans coordinate gaze, head, and body movements in dynamic environments.  

Another line of research incorporates environmental cues for motion prediction. Gaze-informed full-body forecasting has been explored in navigation tasks, using gaze to anticipate walking trajectories \cite{kratzer2020mogaze,zheng2022gimo,yan2023gazemodiff,hu2024gazemotion}. However, these tasks are simpler, as locomotion follows scene constraints rather than requiring complex visuomotor coordination. Full-body forecasting in interactive activities integrates environmental affordances to predict human motion \cite{zheng2022gimo,ashutosh2024fiction}, but such methods heavily rely on external cues like object locations or scene semantics, making them less generalizable to unstructured or unseen environments.  

In contrast, our work addresses a more complex and generalizable problem by unifying head orientation, gaze, and upper-body motion as predictive signals for motion forecasting. By focusing primarily on internal visuomotor cues, our model learns a biologically grounded representation applicable across diverse activities, without explicitly relying on structured environmental priors.

\subsection{Bridging Visuomotor and Imitation Learning}
Recent advances in visuomotor imitation learning have enabled robots to acquire complex skills from human demonstrations \cite{finn2017one,zeng2020transporter,wang2021generalization,lynch2019learning,robomimic2021,zhu2023learning,kim2024openvla,black2024pi,aldaco2024aloha,jiang2024dexmimicgen}. Works like EgoMimic \cite{kareer2024egomimic} explore direct learning from human video data, removing the need for explicit robot data collection. However, modeling human visuomotor coordination remains a challenge, as it involves intricate couplings between gaze, head, and body movements that reflect hidden decision-making processes \cite{lin2025sim}.  

To simplify learning, many existing methods instruct demonstrators to behave ``robot-like" by minimizing natural head and body movements \cite{bahl2022human,li2024okami,zhu2024vision,ren2025motion}. While effective for policy learning, this approach loses the richness of natural human behavior, making it difficult to generalize beyond constrained demonstrations.  

Our work contributes to this direction by providing a predictive model of human visuomotor coordination, capturing the structured dependencies between head pose, gaze, and upper-body motion. Unlike imitation learning approaches that focus on end-to-end policy learning, our model predicts natural visuomotor behavior from in-the-wild datasets \cite{grauman2024ego}. By explicitly modeling visuomotor coordination, our framework can serve as a foundation for future imitation learning research, providing data-driven insights into human movement dynamics that can improve robot learning from human demonstrations.

%% file: 03_method.tex
\begin{figure}[t]
  \centering
   \includegraphics[width=0.9\linewidth]{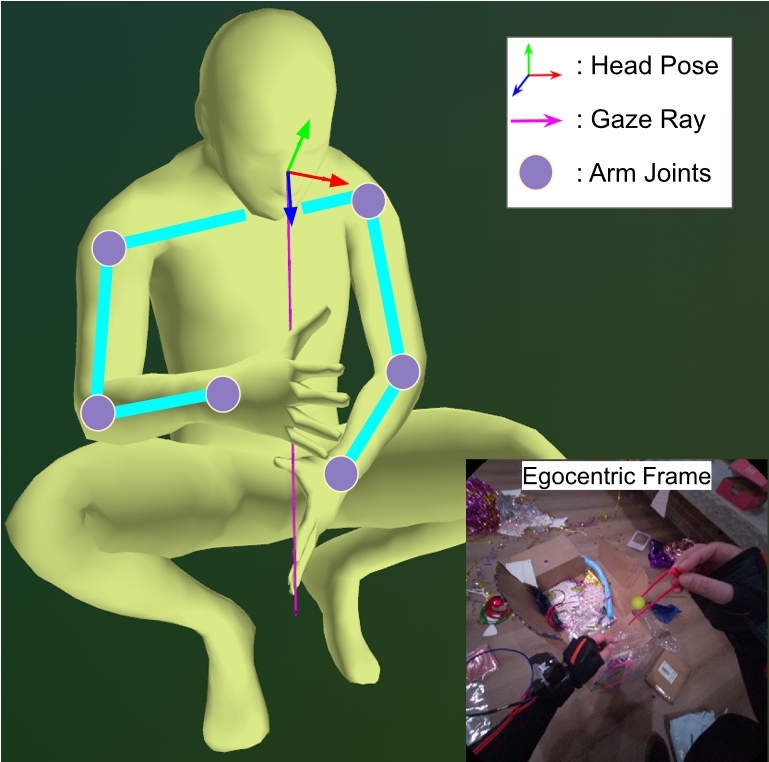}
   \caption{Visualizing the Visuomotor Coordination Representation by mapping it onto a human mesh for better interpretability.}
   \label{fig:visuomotor-rep}
\end{figure}

\section{Method}
\label{sec:method}


We define the \textit{Visuomotor Coordination Representation} (Sec.~\ref{subsec:vcr}) and introduce the forecasting task (Sec.~\ref{subsec:forecasttask}). To remove absolute head motion and ensure temporal and spatial consistency, we apply a state canonicalization process (Sec.~\ref{subsec:canonicalization}). We present our diffusion-based generative framework for visuomotor learning in (Sec.~\ref{subsec:arch}).

\subsection{Human Visuomotor Representation}
\label{subsec:vcr}
To effectively capture human visuomotor coordination, we consider three essential components: the \textit{head pose}, which establishes a spatial reference frame for movement; the \textit{gaze point}, which serves as an indicator of visual attention and intent; and the \textit{upper-body joints}, which provide motion cues relevant to interaction and task execution. These components collectively define the \textit{Visuomotor Coordination Representation (VCR)} by encapsulating the key factors that drive human visuomotor behavior.


We formally define a visuomotor state: \( S = \{ H, G, U \} \), where \( H = (\mathbf{p}_{head}, \mathbf{R}_{head}) \) represents the head pose with position \( \mathbf{p}_{head} \in \mathbb{R}^3 \) and orientation \( \mathbf{R}_{head} \in SO(3) \). The gaze component \( G \) is represented by the \textit{gaze endpoint} \( \mathbf{g} \), defined as \(\mathbf{g} = \mathbf{p}_{head} + \lambda \mathbf{d}_{gaze}\), 
where \( \mathbf{d}_{gaze} \in \mathbb{R}^3 \) is the unit gaze direction derived from \( \mathbf{R}_{head} \), and \( \lambda \) is a predefined scalar controlling the gaze ray length. Finally, \( U = \{ \mathbf{j}_i \in \mathbb{R}^3 \mid i = 1, \dots, 6 \} \) represents the upper-body joint positions, including shoulders, elbows, and wrists. \cref{fig:visuomotor-rep} visualizes an example of VCR from Nymeria~\cite{ma2024nymeria}.

\subsection{Predicting Visuomotor Coordination}
\label{subsec:forecasttask}
Given a Visuomotor Coordination Representation (S) sequence \( S_{t-\tau:t} = \{S_t, S_{t-1}, \dots, S_{t-\tau}\} \) and its corresponding egocentric video clip \( E \in \mathbb{R}^{C \times T \times H \times W} \) , our goal is to predict the future visuomotor states \( \hat{S}_{t+1:t+\Delta} \) over a prediction horizon of \( \Delta \) steps. This task requires joint modeling of the temporal evolution of human motion, including head pose, gaze, and upper-body joint dynamics, based on past visual and kinematic observations. 

Formally, we define the \textit{Visuomotor Forecasting Task} as learning a function:
\[
\hat{S}_{t+1:t+\Delta} = f(S_{t-\Delta:t}, E_{t-\Delta:t}),
\]
where \( f(\cdot) \) models the transition dynamics of visuomotor states given historical motion and egocentric perception, from which enables anticipation of human visuomotor coordination in dynamic environments.

\begin{figure}[t]
  \centering
   \includegraphics[width=0.9\linewidth]{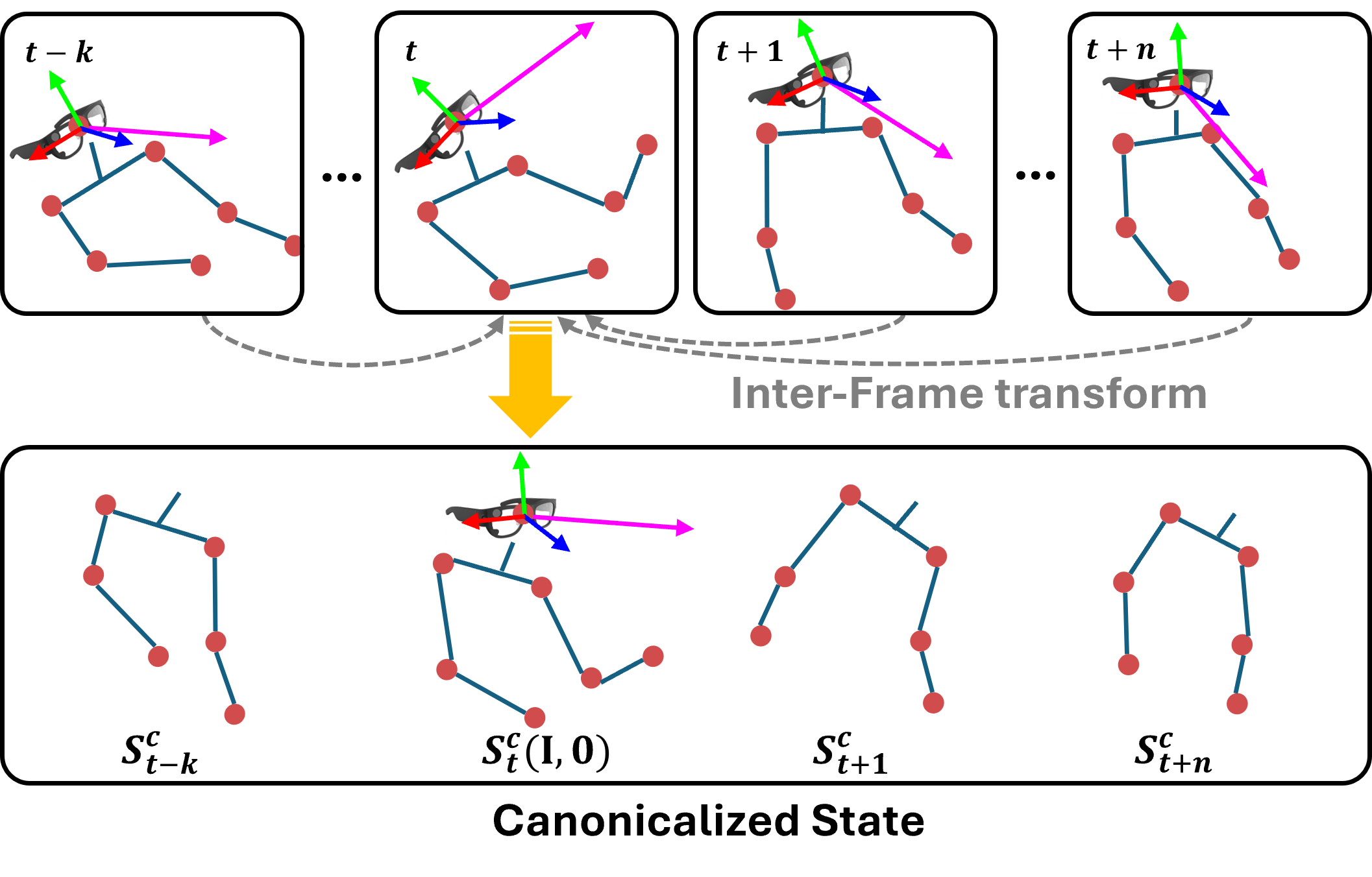}
    \caption{Illustration of the canonicalization process for visuomotor states that help mitigate the effects of absolute head motion and viewpoint variations.}
   \label{fig:visuomotor-canon}
   
\end{figure}
\subsection{Visuomotor State Canonicalization} 
\label{subsec:canonicalization}
We canonicalize kinematic data to ensure consistency across time and spatial frames. To achieve this, we standardize all motion data by defining a stable reference frame based on the final observation. Specifically, we set the head pose \( H_t = (\mathbf{p}_{\text{head},t}, \mathbf{R}_{\text{head},t}) \) in the \textit{last observed step} to a canonical state, applying the transformation \( H^c_t = \Phi(H_t) \), where \( \Phi \) normalizes the head pose by aligning its orientation to identity rotation \( \mathbf{I} \) and positioning it at the origin \( \mathbf{0} \). This transformation ensures consistency across time, allowing the model to learn visuomotor coordination independently of absolute head motion.  

To preserve intra-state spatial consistency, the same transformation \( \Phi \) is applied to the gaze endpoint \( \mathbf{g}_t \) and upper-body joint positions \( U_t \) through \( \mathbf{g}^c_t = \Phi(\mathbf{g}_t) \) and \( U^c_t = \Phi(U_t) \). This results in \( S^c_t = \Phi(S_t)\), where all visuomotor elements remain consistent relative to the canonical head frame, ensuring the model captures meaningful coordination patterns without being affected by head motion variations.  


To preserve inter-state temporal consistency, we extend this operation to all preceding and future kinematic states. For kinematic states \( S_i \) with \( i \in [t - \tau, t + \Delta] \), each state is first transformed relative to \( S_t \) before being mapped to the canonical frame. As a result, we have  
\(
S^c_i = T_{i \to t}(S_i) \circ S^c_t,
\)
where \( S^c_t = \Phi(S_t) \) is the canonicalized reference state. This ensures that all frames remain correctly aligned relative to each other, preserving both temporal and spatial consistency.

This process removes absolute head motion while preserving the relative relationships between the head, gaze, and upper-body joints. It enables the model to learn visuomotor coordination independently of viewpoint variations, improving robustness and generalization. An example of this processing pipeline is shown in Fig.~\ref{fig:visuomotor-canon}. Note that \( S^c_t \) is used in the experiments; however, for simplicity, we continue using the notation \( S_t \) throughout this paper.

\subsection{Model Architecture}
\label{subsec:arch}
\begin{figure}[t]
  \centering
   \includegraphics[width=1.0\linewidth]{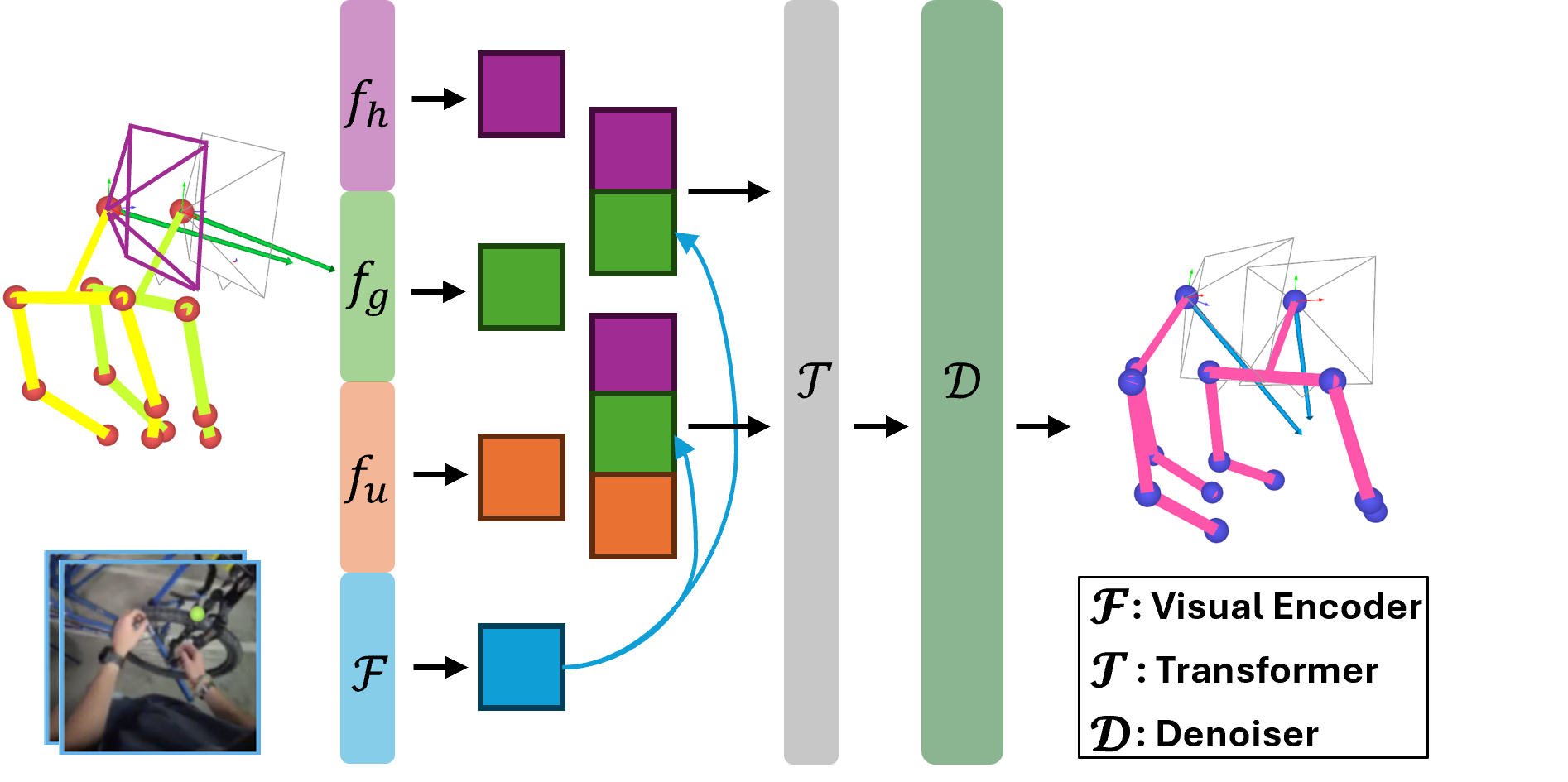}
   \caption{Model Architecture.}
   \label{fig:modelarch}
\end{figure}

We first describe how we extract and fuse multimodal information to construct the conditioning feature for diffusion (Sec.~\ref{subsec:cond_feat}), followed by the diffusion-based visuomotor prediction process (Sec.~\ref{subsec:diffusion_pred}). Figure~\ref{fig:modelarch} provides an overview of our model.
\subsubsection{Conditioning Feature Extraction}
\label{subsec:cond_feat}

\noindent \textbf{Multimodal Feature Encoding}  
Given a sequence of visuomotor states \( S_{t-\tau:t} \) sampled at 10 fps, where each state  
\( S_t = \{ H_t, G_t, U_t \} \) consists of head pose, gaze direction, and upper-body joints,  
we project them into a latent space using learned functions:
\(
\mathbf{k}_t^{H} = f_h(H_t), \quad \mathbf{k}_t^{G} = f_g(G_t), \quad \mathbf{k}_t^{U} = f_u(U_t).
\)
Egocentric RGB frames \( E_{t-\tau:t} \) are sampled at 4 fps, providing complementary contextual information  
about the environment and task-relevant objects. A single visual embedding  
\(
\mathbf{v} = \mathcal{F}_{\text{vis}}(E_{t-\tau:t}) \in \mathbb{R}^{128}
\)
is extracted from the sequence using a 3D ResNet backbone.

\noindent \textbf{Multimodal Feature Fusion.} While egocentric frames always contain cues directly related to head and gaze orientation, their relevance to full visuomotor coordination is uncertain due to partial occlusions and viewpoint limitations. To mitigate this, instead of using a single kinematic representation, we construct two separate kinematic representations: one capturing head and gaze features, defined as \( \mathbf{k}^{\text{hg}}_t = \text{Concat}(\mathbf{k}^{\text{head}}_t, \mathbf{k}^{\text{gaze}}_t) \), and another incorporating full-body motion as \( \mathbf{k}^{\text{hga}}_t = \text{Concat}(\mathbf{k}^{\text{head}}_t, \mathbf{k}^{\text{gaze}}_t, \mathbf{k}^{\text{arm}}_t) \). 

\noindent We apply cross-attention separately to \( \mathbf{k}^{\text{hg}}_t \), which captures viewpoint and attentional dynamics, and to \( \mathbf{k}^{\text{hga}}_t \), which additionally includes upper-body motion cues. This structured fusion allows the model to selectively incorporate spatial and attentional signals while maintaining robustness to missing or ambiguous visual information. Formally, we compute:
\begin{equation*}
    \begin{aligned}
        \mathbf{k}^{\prime \text{hg}}_t &= \mathcal{A}(\mathbf{k}^{\text{hg}}_t, \mathbf{v}, \mathbf{v}), \quad 
        \mathbf{k}^{\prime \text{hga}}_t &= \mathcal{A}(f_{\text{proj}}(\mathbf{k}^{\text{hga}}_t), \mathbf{v}, \mathbf{v}).
    \end{aligned}
\end{equation*}

\noindent The final fused representation is obtained by summing the attended features:  
\( \mathbf{k}_t^{\text{fused}} = \mathbf{k}^{\prime \text{hg}}_t + \mathbf{k}^{\prime \text{hga}}_t \), which is then passed to a Transformer-based temporal encoder \(\mathcal{T}\) for sequential modeling. The output is flattened into a single conditioning feature \( \mathbf{c} \), which is used as input to the denoiser \(\mathcal{D}\).

\subsubsection{Diffusion-Based Visuomotor Prediction}
\label{subsec:diffusion_pred}
We follow \cite{chi2023diffusion} to formulate visuomotor forecasting as a denoising diffusion process, where the model learns to iteratively refine a noisy sequence into a future trajectory. 

\noindent \textbf{Forward Diffusion Process.}  
Following the standard denoising diffusion probabilistic model (DDPM) \cite{ho2020denoising}, we define the forward process as a Markovian sequence that progressively adds Gaussian noise to the ground-truth future visuomotor states \( S_{t+1:T} \):
\begin{equation*}
    q(S_t | S_0) = \mathcal{N}(S_t; \sqrt{\bar{\alpha}_t} S_0, (1 - \bar{\alpha}_t) \mathbf{I})
\end{equation*}
where \( \bar{\alpha}_t \) is a pre-defined noise schedule. This process converts the original data distribution into an isotropic Gaussian in the latent space.

\noindent \textbf{Reverse Denoising Process.}  
The model learns to iteratively denoise the corrupted future states using a neural network parameterized by \( \theta \). The reverse process is defined as:

\begin{equation*}
    p_{\theta}(S_{t-1} | S_t, \mathbf{c}) = \mathcal{N}(S_{t-1}; \mu_{\theta}(S_t, t, \mathbf{c}), \sigma_{\theta}^2 \mathbf{I})
\end{equation*}

where \( \mu_{\theta} \) is the predicted mean, \( \sigma_{\theta}^2 \) is the variance, and the conditioning feature \( \mathbf{c} \) remains constant throughout the denoising process.

\noindent The model is trained using a standard denoising loss:

\begin{equation*}
    \mathcal{L} = \mathbb{E}_{S_0, t, \boldsymbol{\epsilon}} \left[ || \boldsymbol{\epsilon} - \epsilon_{\theta}(S_t, t, \mathbf{c}) ||^2 \right]
\end{equation*}

where \( \epsilon_{\theta} \) represents the noise prediction network.

\noindent At inference time, we sample future visuomotor trajectories by iteratively applying the learned reverse process, starting from a Gaussian prior.

\subsubsection{Implementation Details}
\noindent \textbf{Training Details.} The visual encoder is pre-trained on KINETICS400\_V1, while both the transformer module and the DDPM diffusion model are trained from scratch. The model is implemented in PyTorch and trained for 400 epochs using the AdamW optimizer with a learning rate of \( 5 \times 10^{-4} \). Training is conducted on a single H100 GPU, with a batch size of 384, requiring approximately 8 hours to complete.

%% file: 04_experiment.tex
\section{Experiments}
\label{sec:experiments}
We conduct extensive experiments to evaluate our predictive visuomotor learning framework. We first introduce the datasets used in our study (Sec.~\ref{subsec:dataset}), followed by the evaluation metrics (Sec.~\ref{subsec:metrics}) used to assess prediction accuracy. We then present the baselines for comparison (Sec.~\ref{subsec:baseline}) and analyze the impact of different model components through ablation studies (Sec.~\ref{subsec:abla}). Finally, we provide both qualitative (Sec.~\ref{subsec:quali}) and quantitative (Sec.~\ref{subsec:quanti}) results to demonstrate the effectiveness of our approach.  

\subsection{Datasets}
\label{subsec:dataset}

\noindent \textbf{EgoExo4D} \cite{grauman2024ego} is a large-scale multimodal egocentric-exocentric video dataset collected by the Ego4D Consortium. It utilizes Aria glasses \cite{engel2023project} alongside four GoPro cameras to capture aligned first-person (egocentric) and third-person (exocentric) views of skilled activities across more than 130 scene contexts, totaling approximately 88 hours of egocentric footage.  
The dataset provides head pose and gaze annotations computed by Meta’s Multimodal Perception Service (MPS), ensuring accurate SLAM-based pose estimation. Additionally, full-body 3D joint positions are annotated from the exocentric videos. While EgoExo4D encompasses a diverse range of activities, not all are equally suited for studying visuomotor coordination. We select a subset of activities where participants manipulate objects and navigate their environment in ways that naturally integrate visual attention with motor actions. The selected scenarios include Basketball, Cooking, Bike Fixing, and Health-related tasks, resulting in 23,372 training samples and 5,126 testing samples, with a total duration of approximately 15.8 hours. Detailed information about the data cleaning pipeline and per-class statistical distribution is available in Sec.~\ref{sec:dataclean} of our supplementary materials.

\subsection{Evaluation Metrics}
\label{subsec:metrics}
To comprehensively evaluate our predictive visuomotor learning task, we adopt a diverse set of metrics that assess structural consistency, positional accuracy, and orientation alignment. These metrics are computed over the visuomotor representation \( S = \{ H, G, U \} \), as defined in Sec.~\ref{subsec:vcr}.

\noindent\textbf{Structural Consistency.} \textbf{PA-MPJPE} (mm) measures the structural consistency of predicted visuomotor coordination. Unlike absolute error metrics, PA-MPJPE aligns the predicted and ground-truth poses via a rigid transformation, ensuring that the error primarily reflects internal joint structure deviations rather than global displacement. This metric is computed over \( \{\mathbf{p}_{head}, G, U\} \).

\noindent\textbf{Position Accuracy.} We measure the Euclidean distance between the predicted and ground-truth positions of key body parts. Specifically, \textbf{Head, Gaze, and Hand Errors} (mm) correspond to \( \mathbf{p}_{head} \), \( G \), and the wrist joints \( \mathbf{j}_i \) for \( i \in \{5,6\} \), which are a subset of \( U \), respectively.

\noindent\textbf{Orientation Accuracy.} We evaluate the accuracy of head rotation with \textbf{Head Rotation Error (HRE)} (degree) measures the angular deviation between the predicted and ground-truth head orientations.


\subsection{Baselines}
\label{subsec:baseline}

To evaluate the effectiveness of our method, we compare it against two naïve interpolation baselines and two learning-based methods, including a Diffusion Policy model.\footnote{Our work is concurrent with EgoCast~\cite{escobar2025egocast} and EgoAgent~\cite{chen2025acquisition}, but a direct comparison is not feasible due to differences in task definitions and input modalities, as well as the lack of publicly available implementations at the time of writing.}

\begin{itemize}
    \item \textbf{Constant Pose}: assumes no future motion, directly copying the last observed visuomotor state for all future time steps. It serves as a lower bound, representing a scenario where no predictive modeling is performed.
    \item \textbf{Constant Velocity}: estimates future motion using first-order linear extrapolation, assuming a constant velocity based on the last observed state transition. It provides a simple motion forecasting heuristic.
    \item \textbf{Transformer Model}: is a standard Transformer-based sequence-to-sequence predictor that models motion evolution through self-attention. We concatenate pose sequences with visual embeddings extracted from ResNet18 and feed them into a 3-layer Transformer encoder, followed by a regression head to predict future visuomotor states. 
    \item \textbf{Diffusion Policy (CNN-Based) \cite{chi2023diffusion}}: formulates motion prediction as a denoising diffusion process, where a neural network iteratively refines a noisy motion sequence into a plausible future trajectory. We apply this framework to our setting by extracting egocentric frame features using a pre-trained ResNet18, concatenating them with raw visuomotor observations, and flattening the resulting feature representation to condition a U-Net denoiser. 
\end{itemize}

\renewcommand{\arraystretch}{1.1}  
\begin{table*}[h!]
\centering
\footnotesize 
{\setlength{\tabcolsep}{8pt} 
    \begin{tabular}{|c|c|c|c|c|c|}
        \hline
         Methods & PA-MPJPE, \% & Head Pos., \% & Gaze Pos., \% & Hand Pos., \% & Head Rot., \%\\ \hline
        \textsc{Constant Pose} &  68.3, +16.6 &  184, +73.6 & 193, +55.6 & 274, +45.7 & 16.7, +25.6 \\ 
        \textsc{Constant Velocity} &  109, +84.7 & 161, +51.9 & 201, + 62.1 & 436, +132  &  18.5, +39.1 \\ 
        Transformer Encoder + MLP &  65.3, +10.7 & 119, +12.3 & 135, +8.1 & 211, +12.2 & 13.8, +4.5 \\
        Diffusion Policy-CNN \cite{chi2023diffusion} & 64.1, +8.6  & 112, +5.7 & 132, +6.5 & 208, +10.6 & 13.9, +5.3 \\
        \cellcolor{green!20}\textbf{Ours} & \cellcolor{green!20}\textbf{59}  & \cellcolor{green!20}\textbf{106} & \cellcolor{green!20}\textbf{124} & \cellcolor{green!20}\textbf{188} & \cellcolor{green!20}\textbf{13.2} \\ \hline
    \end{tabular}}
    \caption{Quantitative comparison of our method against baselines. Lower values indicate better performance.
    }
\label{tab:baseline}
\end{table*}

\subsection{Ablations}
\label{subsec:abla}
To assess the contribution of different components in our visuomotor prediction model, we conduct two types of ablations: one focusing on specific input and output signals, and another analyzing the impact of different modalities.

\noindent \textbf{Input-Output Ablations.} 
We first examine the role of key visuomotor signals by selectively removing them from the input or output space:

\begin{itemize}
    \item \textbf{w/o Head Rotation}: Removes head rotation from the output and optionally from the input to test its necessity.
    \item \textbf{w/o Head Rotation \& Gaze}: Removes both head rotation and gaze from the output and optionally from the input to evaluate their impact.
    \item \textbf{w/o Head}: Removes all head-related information (position and rotation) from the output and optionally from the input.
    \item \textbf{w/o Gaze}: Removes gaze from the output and optionally from the input to test its role.
\end{itemize}

\noindent \textbf{Signal Modality Ablations.} 
We further analyze the contribution of different sensory and temporal modalities:

\begin{itemize}
    \item \textbf{w Last Step Arm}: Uses only the last observed arm configuration instead of the full motion history to evaluate the importance of temporal context.
    \item \textbf{w/o Egocentric Frame}: Removes egocentric visual input, leaving only kinematic information, to assess the role of first-person vision in visuomotor prediction.
\end{itemize}

\begin{table*}[htbp]
\centering
\footnotesize 
{\setlength{\tabcolsep}{8pt} 
    \begin{tabular}{|c|c|c|c|c|c|c|}
        \hline
         Ablations & Input & Head Pos., \% & Gaze Pos., \% & Hand Pos., \% & Head Rot., \% & PA-MPJPE, \% \\ \hline
        \rowcolor{green!20}\textbf{}
        Complete Visuomotor  & - & \textbf{106} & \textbf{124} & \textbf{188} & \textbf{13.2} & \textbf{59}\\ \hline
        \multicolumn{7}{|c|}{\cellcolor{yellow!20}\textbf{Input-Output Ablations}} \\ \hline
        Head Rotation  & \bluecheck & 108, +1.9 & 126, +1.6 & 188, +0.0 & {\color{red} \textbf{\xmark}} & \multirow{8}{*}{\centering \parbox{3cm}{\centering \textit{Metric unavailable due to incomplete output space}}} \\
        Head Rotation  & {\color{red} \textbf{\xmark}}  & 111, +4.7 & 130, +4.8 & 195, +3.7 & {\color{red} \textbf{\xmark}} &  \\ \cline{1-6}
        Head Rotation \& Gaze  & \bluecheck & 109, +2.8 &  {\color{red} \textbf{\xmark}}  & 190, +1.1 & {\color{red} \textbf{\xmark}} &  \\
        Head Rotation \& Gaze  & {\color{red} \textbf{\xmark}} & 112, +5.7 & {\color{red} \textbf{\xmark}} & 196, +4.3 & {\color{red} \textbf{\xmark}} & \\ \cline{1-6}
        Head  & \bluecheck & {\color{red} \textbf{\xmark}}  & 127, +2.4 & 190, +1.1 & {\color{red} \textbf{\xmark}} &  \\
        Head  & {\color{red} \textbf{\xmark}} & {\color{red} \textbf{\xmark}} & 132, +6.5 & 194, +3.2 & {\color{red} \textbf{\xmark}} &  \\ \cline{1-6}
        Gaze  & \bluecheck & 109, +2.8  & {\color{red} \textbf{\xmark}} & 190, +1.1 & 13.4, +1.5 &  \\ 
        Gaze  & {\color{red} \textbf{\xmark}} & 111, +4.7 & {\color{red} \textbf{\xmark}}  & 194, +3.2 & 13.9, + 4.5 &  \\ 
        \hline 
        \multicolumn{7}{|c|}{\cellcolor{yellow!20}\textbf{Signal Modality Ablations}} \\ \hline
        w Last Step Arm & - & 113, + 6.6 & 141, + 5.2 & 199, + 5.9 & 13.7 , + 3.8  & 61, + 3.4 \\
        w/o Egocentric Frame & - & 111, + 4.7 & 130, + 4.8 & 193, + 2.7 & 14.1 , + 6.0 &  60, + 1.7 \\
        \hline 
    \end{tabular}}
    \caption{\textbf{Ablation Study on Input Configuration and Signal Modalities.} The table presents two sets of ablations: \textit{(1) Input-Output Ablations}, which evaluate the impact of removing specific kinematic inputs (head rotation, head position, gaze, and upper-body joints) on prediction accuracy, and \textit{(2) Signal Modality Ablations}, which assess the effect of removing higher-level input signals such as the last-step arm pose and egocentric frames. The \textit{Complete Visuomotor} row represents the full model, achieving the lowest error across all metrics.}
\label{tab:ablation}
\end{table*}


\subsection{Quantitative Results}
\label{subsec:quanti}
Tables~\ref{tab:baseline} and \ref{tab:ablation} summarize our results: baseline comparisons, ablation studies on different input modalities, and per-class prediction performance, respectively. The best results are highlighted in bold, and lower values across all metrics indicate better performance (Sec.~\ref{subsec:dataset}).

\noindent \textbf{Comparison with Baseline Methods.}
Table~\ref{tab:baseline} compares our method with four baselines for visuomotor prediction. We first examine naïve interpolation baselines, Constant Pose and Constant Velocity, which assume either no motion or simple extrapolation. As expected, these baselines perform poorly. Constant Pose yields a PA-MPJPE of 68.3 and a hand error of 274, showing the inadequacy of static predictions. Constant Velocity performs even worse, with PA-MPJPE reaching 264 and hand error rising to 436, indicating that simple extrapolation is insufficient for modeling complex visuomotor behavior.

Learning-based approaches significantly outperform interpolation baselines, demonstrating the need for temporal modeling and multimodal integration. Among prior works, Diffusion Policy-CNN and Transformer Encoder + MLP serve as strong baselines. Our model achieves a PA-MPJPE of 59, improving upon Diffusion Policy-CNN by 8.6\% and the Transformer-based baseline by 10.7\%. Similarly, all of our position errors outperform them, and our model improves the head and gaze error of Diffusion Policy by 5.7\% and 6.5\%, respectively. These improvements indicate that our method generates more precise motion predictions while maintaining realistic visuomotor coordination.

Notably, our model achieves the largest improvement in predicting hand position, which is the most challenging sub-task. This suggests that our approach effectively captures head-eye-hand coordination, leveraging structured visuomotor representations for fine-grained motion prediction.

When evaluating the orientations of head and gaze, our model achieves 13.2 degree as the average error across all steps, improving upon Diffusion Policy-CNN by 4.5\% and the Transformer method by 6\%. The fact that our model performs well across both head rotation (HRE) and translation suggests that it effectively learns a unified visuomotor representation, handling different motion modalities within the same framework.

\noindent \textbf{Ablation Study.}  
Table~\ref{tab:ablation} evaluates the impact of removing different input components on visuomotor prediction. The \textit{Complete Visuomotor} setting (row 0) includes all available inputs (head pose, gaze, and upper-body joints), while the ablations systematically remove each to examine its role.  

Removing head rotation (row 2) increases head position error from 108 to 111, while additionally removing gaze input (row 4) further degrades performance to 112. Gaze position error rises from 126 to 130, and hand position error increases from 188 to 196, highlighting the importance of head and gaze signals for accurate motion prediction. When all head-related inputs are removed (row 6), gaze error increases from 127 to 132, and hand error from 190 to 194, indicating that head motion affects overall visuomotor coordination. The absence of gaze input (row 8) also increases head rotation error from 13.4 to 13.9, confirming its role in stabilizing head orientation.  

The lower section of Table~\ref{tab:ablation} examines high-level signal ablations. Preserving only the last-step arm pose (\textit{w/ Last Step Arm}) increases head, gaze, and hand errors by 6.6\%, 5.2\%, and 5.9\%, respectively, suggesting that a single arm pose lacks sufficient temporal context for predicting upper-body motion. Removing egocentric vision (\textit{w/o Egocentric Frame}) leads to 4.7\% higher head position error and a 6\% increase in head rotation error, reinforcing the role of visual context in stabilizing head and gaze coordination. Interestingly, gaze error decreases slightly (124 to 130), suggesting a shift toward kinematic reliance for gaze estimation.  

These results confirm that multimodal integration enhances predictive visuomotor coordination: temporal kinematic history improves hand predictions, while egocentric vision stabilizes head and gaze alignment.

\begin{figure*}[t]
  \centering
   \includegraphics[width=1.0\linewidth]{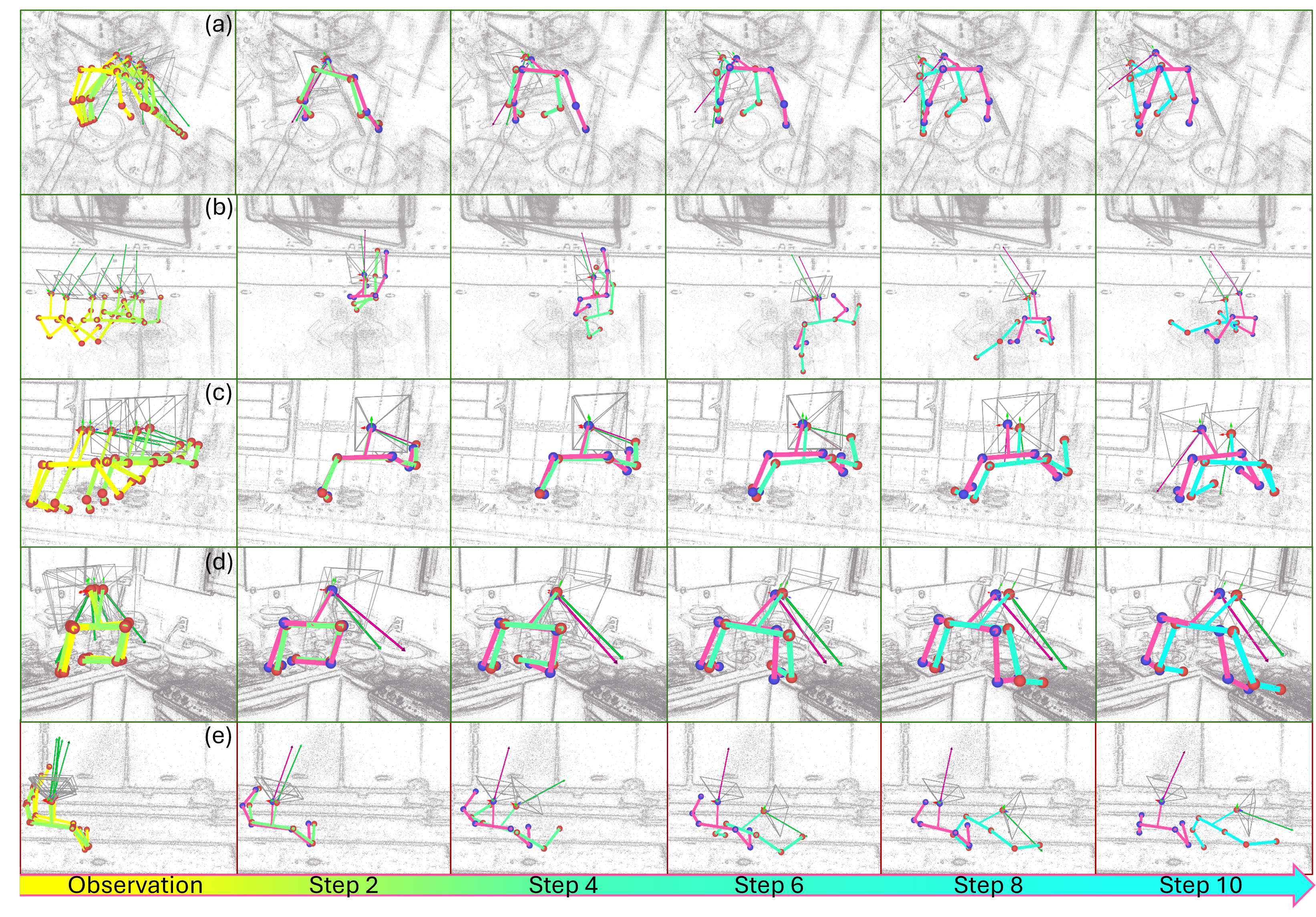}
    \caption{Visualization of predicted visuomotor coordination across diverse real-world scenes. Each row corresponds to a different scene, while columns illustrate the temporal evolution of visuomotor coordination. The {\color{yellow}{\textbf{first column}}} represents the observed states, while subsequent columns display predictions from time step 2 to 10. {\color{magenta}{\textbf{Predicted poses}}} and {\color{cyan}{\textbf{ground truth poses}}} are overlaid for interpretability. View in color for best results.}
   \label{fig:qualitative}
\end{figure*}

\subsection{Qualitative Results}  
\label{subsec:quali}

Figure~\ref{fig:qualitative} illustrates the predicted visuomotor coordination across diverse real-world scenes. The model effectively captures head, gaze, and upper-body dynamics, maintaining smooth and temporally consistent motion. As shown in Rows (a)-(d), it successfully anticipates head and gaze shifts, even when adapting to scene constraints, demonstrating its ability to infer visuomotor intent from prior observations.


However, failure cases occur in scenarios with rapid, unexpected movements or occlusions, where subtle cues are insufficient for precise coordination. While our model still produces reasonable predictions, it may misalign with the ground truth. Row (e) presents a typical failure case: as the basketball bounces quickly—a nuance observable only in the last frame of the egocentric input—the person rapidly shifts their body rightward to catch it. In contrast, our prediction assumes a standard catching motion, failing to adapt to the sudden trajectory change. Despite these challenges, overall prediction quality remains strong, reinforcing the effectiveness of multimodal integration in learning visuomotor coordination. Future improvements could involve incorporating explicit contact modeling or leveraging environment-aware reasoning to enhance robustness in highly dynamic tasks. For a more comprehensive demonstration, we provide video examples in the supplementary materials.

%% file: 10_conclusion.tex
\section{Conclusion}  
We introduced a forecasting-based task for human visuomotor modeling, where the goal is to predict future head pose, gaze, and upper-body motion from egocentric visual and kinematic observations. To achieve this, we proposed the \textit{Visuomotor Coordination Representation}, which learns structured temporal dependencies across multimodal signals. We further extended a diffusion-based motion modeling framework integrating egocentric vision and kinematic sequences, enabling temporally coherent and accurate visuomotor predictions. Our approach was evaluated on the challenging large-scale EgoExo4D dataset, demonstrating strong generalization across diverse real-world activities. Through extensive experiments, we showed that multimodal integration plays a crucial role in improving visuomotor modeling. Our results provide insights into structured human motion representation, contributing to applications in robotics, human-computer interaction, and assistive technologies.

%% file: 12_appendix.tex
\section{Additional Video Demo}
\label{sec:qual_results}
Due to space limitations, the main paper only presents static visualizations. In this supplementary material, we provide a richer set of qualitative demos showcasing our model’s predictions across diverse scenarios. These include extended visualizations of predicted head pose, gaze, and upper-body motion, highlighting both successful cases and failure modes. We also analyze common prediction errors, such as subtle signal leads to wrong predictions, and inherent challenge from unforseeable human motion, to better illustrate the model’s strengths and limitations.  

We encourage readers to view the full set of qualitative results in the provided demo videos for a more comprehensive understanding of our model’s performance.

\section{EgoExo4D Data Cleaning Pipeline}
\label{sec:dataclean}
EgoExo4D provides detailed pose annotations by running off-the-shelf human pose estimation models on multiple exocentric camera views. However, occlusions frequently lead to missing body parts, resulting in occasional inaccuracies in the automatically generated annotations. Even with manually annotated corrections, these issues remain common due to unavoidable viewpoint limitations.  

While Fiction~\cite{ashutosh2024fiction} improves annotation quality by re-annotating filtered sequences using narration-based semantic cues, our goal is to model general visuomotor coordination without restricting the dataset based on activity type. Instead of manually filtering data based on semantics, we apply a 5-second sliding window to correct annotation errors using temporally adjacent frames. If a missing or incorrect joint annotation cannot be recovered within a reasonable range, we discard that frame.  

After this cleaning process, our training and testing samples are drawn from the valid index list using a 20 steps sliding window, with a stride of 10 steps. This ensures that our model learns from reliable annotations while maintaining a broad range of natural visuomotor behaviors, aligning with our goal of capturing general coordination patterns rather than task-specific motions.

\begin{table}[h!]
\centering
\footnotesize 
{\setlength{\tabcolsep}{8pt} 
    \begin{tabular}{|c|c|c|c|c|}
        \hline
        Task / \( \Delta_{Avg} \) &  Basketball & Cooking & Bike & Health \\ \hline
        PA-MPJPE & 78 / 116  & 47 / 59 & 52 / 61 & 38 / 48 \\ 
        Head Pos. &  16 / 35  & 12 / 18 & 11 / 19  & 11 / 17 \\
        Gaze Pos.  &  195 / 546 & 89 / 164 & 85 / 137 & 64 / 94 \\
        Hand Pos.  & 304 / 733  & 128 / 208 & 124 / 166 & 87 / 117 \\ 
        Head Rot. &  16 / 35  & 12 / 18 & 11 / 19  & 11 / 17 \\ \hline
        Count &  2,037 & 887 & 1,136 & 1,066 \\ \hline
    \end{tabular}}
    \caption{Per-class Average Prediction Error vs. Motion Change Amplitude across different activity categories.}
    \label{tab:perclass}
\end{table}

\begin{table}[h!]
\centering
\footnotesize 
{\setlength{\tabcolsep}{8pt} 
    \begin{tabular}{|c|c|c|c|c|c|c|}
        \hline
        Time Step  & t+1 & t+3 & t+5 & t+7 & t+10 & Mean \\ \hline
        PA-MPJPE   & 29  & 48 & 60 & 68 & 78 &  59 \\ 
        Head Pos. & 16 & 54 & 94 & 136 & 200 & 106 \\
        Gaze Pos. & 26 & 69 & 112 & 156 & 226 & 124 \\
        Hand Pos.  & 61 & 130 & 181  & 228 & 294 & 188 \\ 
        Head Rot. & 2.5 & 7.6 & 12.3 & 16.7 & 23.2 & 13.2 \\ \hline
    \end{tabular}}
    \caption{Per-Step Performance vs. Mean Performance.}
    \label{tab:perstep}
\end{table}

\section{Per-Class Performance Analysis}  
Table~\ref{tab:perclass} presents the per-step prediction performance of our model across different skilled activities. The results reveal a strong correlation between motion variability and prediction difficulty, with activities exhibiting larger motion change amplitudes (\( \Delta_{Avg} \)) leading to higher errors. Structured tasks like Cooking and Health yield lower errors, while dynamic or fine-grained activities such as Basketball and fixing Bike introduce more uncertainty due to abrupt gaze shifts and complex hand-eye coordination. These findings highlight the challenge of modeling high-motion scenarios, suggesting that improving robustness in such conditions is crucial for advancing visuomotor prediction models.

\section{Per-Step Performance Analysis.}
Table \ref{tab:perstep} presents the per-step prediction performance of our model across different future time steps. While the diffusion model generates the entire trajectory at once, errors increase over longer horizons due to growing uncertainty in future motion. At \( t+1 \), predictions are highly accurate, with head position error at 16 and head rotation at 2.5 degrees. However, as the time step extends, errors grow significantly, reaching 200 for head position and 226 for gaze position at \( t+10 \), reflecting the increasing difficulty of modeling long-range dependencies where future states become less constrained by recent observations.  

Hand motion exhibits the largest variation, with error rising from 61 at \( t+1 \) to 294 at \( t+10 \), suggesting that fine-grained hand movements are harder to predict due to their higher variability and dependence on external factors. In contrast, head rotation remains more stable, increasing gradually to 23.2 at \( t+10 \), indicating that head orientation follows smoother, more predictable patterns. These results suggest that incorporating trajectory-level constraints or enhancing long-range temporal dependencies could improve long-horizon stability, particularly for hand motion.

%% file: _main.bbl
\begin{thebibliography}{59}
\providecommand{\natexlab}[1]{#1}
\providecommand{\url}[1]{\texttt{#1}}
\expandafter\ifx\csname urlstyle\endcsname\relax
  \providecommand{\doi}[1]{doi: #1}\else
  \providecommand{\doi}{doi: \begingroup \urlstyle{rm}\Url}\fi

\bibitem[Aldaco et~al.(2024)Aldaco, Armstrong, Baruch, Bingham, Chan, Draper, Dwibedi, Finn, Florence, Goodrich, et~al.]{aldaco2024aloha}
Jorge Aldaco, Travis Armstrong, Robert Baruch, Jeff Bingham, Sanky Chan, Kenneth Draper, Debidatta Dwibedi, Chelsea Finn, Pete Florence, Spencer Goodrich, et~al.
\newblock Aloha 2: An enhanced low-cost hardware for bimanual teleoperation.
\newblock \emph{arXiv preprint arXiv:2405.02292}, 2024.

\bibitem[Aronson et~al.(2021)]{Aronson2021}
Robert~M. Aronson et~al.
\newblock Inferring goals with gaze during teleoperated manipulation.
\newblock In \emph{IEEE International Conference on Robotics and Automation (ICRA ’21)}, 2021.

\bibitem[Ashutosh et~al.(2024)Ashutosh, Pavlakos, and Grauman]{ashutosh2024fiction}
Kumar Ashutosh, Georgios Pavlakos, and Kristen Grauman.
\newblock Fiction: 4d future interaction prediction from video.
\newblock \emph{arXiv preprint arXiv:2412.00932}, 2024.

\bibitem[Avraham et~al.(2019)Avraham, Sulimani, Mussa-Ivaldi, and Nisky]{avraham2019effects}
Guy Avraham, Erez Sulimani, Ferdinando~A Mussa-Ivaldi, and Ilana Nisky.
\newblock Effects of visuomotor delays on the control of movement and on perceptual localization in the presence and absence of visual targets.
\newblock \emph{Journal of neurophysiology}, 122\penalty0 (6):\penalty0 2259--2271, 2019.

\bibitem[Bahl et~al.(2022)Bahl, Gupta, and Pathak]{bahl2022human}
Shikhar Bahl, Abhinav Gupta, and Deepak Pathak.
\newblock Human-to-robot imitation in the wild.
\newblock \emph{arXiv preprint arXiv:2207.09450}, 2022.

\bibitem[Bisogni et~al.(2024)]{Bisogni2024}
C. Bisogni et~al.
\newblock Gaze analysis: A survey on its applications.
\newblock \emph{Computers in Human Behavior}, pages 1--18, 2024.

\bibitem[Bizzi et~al.(1984)Bizzi, Accornero, Chapple, and Hogan]{bizzi1984posture}
Emilio Bizzi, Neri Accornero, William Chapple, and Neville Hogan.
\newblock Posture control and trajectory formation during arm movement.
\newblock \emph{Journal of Neuroscience}, 4\penalty0 (11):\penalty0 2738--2744, 1984.

\bibitem[Black et~al.(2024)Black, Brown, Driess, Esmail, Equi, Finn, Fusai, Groom, Hausman, Ichter, et~al.]{black2024pi}
Kevin Black, Noah Brown, Danny Driess, Adnan Esmail, Michael Equi, Chelsea Finn, Niccolo Fusai, Lachy Groom, Karol Hausman, Brian Ichter, et~al.
\newblock pi\_0: A vision-language-action flow model for general robot control.
\newblock \emph{Robotics: Science and Systems XXI}, 2024.

\bibitem[Chen et~al.(2025)Chen, Wang, Tang, Ma, He, Ouyang, Zhou, Bao, and Peng]{chen2025egoagent}
Lu Chen, Yizhou Wang, Shixiang Tang, Qianhong Ma, Tong He, Wanli Ouyang, Xiaowei Zhou, Hujun Bao, and Sida Peng.
\newblock Egoagent: A joint predictive agent model in egocentric worlds.
\newblock 2025.

\bibitem[Chen et~al.(2022)]{Chen2022}
X.~L. Chen et~al.
\newblock Gaze-based interaction intention recognition in virtual reality.
\newblock \emph{Electronics}, 11\penalty0 (10):\penalty0 1647, 2022.

\bibitem[Chi et~al.(2023)Chi, Xu, Feng, Cousineau, Du, Burchfiel, Tedrake, and Song]{chi2023diffusion}
Cheng Chi, Zhenjia Xu, Siyuan Feng, Eric Cousineau, Yilun Du, Benjamin Burchfiel, Russ Tedrake, and Shuran Song.
\newblock Diffusion policy: Visuomotor policy learning via action diffusion.
\newblock \emph{The International Journal of Robotics Research}, page 02783649241273668, 2023.

\bibitem[Chuang et~al.(2025)Chuang, Zou, Lee, Gao, and Soltani]{chuang2025look}
Ian Chuang, Jinyu Zou, Andrew Lee, Dechen Gao, and Iman Soltani.
\newblock Look, focus, act: Efficient and robust robot learning via human gaze and foveated vision transformers.
\newblock \emph{arXiv preprint arXiv:2507.15833}, 2025.

\bibitem[David-John et~al.(2021)David-John, Peacock, Zhang, Murdison, Benko, and Jonker]{DavidJohn2021}
Benjamin David-John, Christian~E. Peacock, Tyler Zhang, Tyler~S. Murdison, Hrvoje Benko, and Tomás~R. Jonker.
\newblock Towards gaze-based prediction of the intent to interact in virtual reality.
\newblock In \emph{Proceedings of the ACM Symposium on Eye Tracking Research \& Applications (ETRA ’21)}, page~7, 2021.

\bibitem[Engel et~al.(2023)Engel, Somasundaram, Goesele, Sun, Gamino, Turner, Talattof, Yuan, Souti, Meredith, et~al.]{engel2023project}
Jakob Engel, Kiran Somasundaram, Michael Goesele, Albert Sun, Alexander Gamino, Andrew Turner, Arjang Talattof, Arnie Yuan, Bilal Souti, Brighid Meredith, et~al.
\newblock Project aria: A new tool for egocentric multi-modal ai research.
\newblock \emph{arXiv preprint arXiv:2308.13561}, 2023.

\bibitem[Escobar et~al.(2025)Escobar, Puentes, Forigua, Pont-Tuset, Maninis, and Arbeláez]{escobar2025egocast}
Maria Escobar, Juanita Puentes, Cristhian Forigua, Jordi Pont-Tuset, Kevis-Kokitsi Maninis, and Pablo Arbeláez.
\newblock Egocast: Forecasting egocentric human pose in the wild.
\newblock \emph{Proceedings of the IEEE/CVF Winter Conference on Applications of Computer Vision}, 2025.

\bibitem[Finn et~al.(2017)Finn, Yu, Zhang, Abbeel, and Levine]{finn2017one}
Chelsea Finn, Tianhe Yu, Tianhao Zhang, Pieter Abbeel, and Sergey Levine.
\newblock One-shot visual imitation learning via meta-learning.
\newblock In \emph{Conference on robot learning}, pages 357--368. PMLR, 2017.

\bibitem[Fragkiadaki et~al.(2015)Fragkiadaki, Levine, Felsen, and Malik]{fragkiadaki2015recurrent}
Katerina Fragkiadaki, Sergey Levine, Panna Felsen, and Jitendra Malik.
\newblock Recurrent network models for human dynamics.
\newblock In \emph{Proceedings of the IEEE international conference on computer vision}, pages 4346--4354, 2015.

\bibitem[Grauman et~al.(2022)Grauman, Westbury, Byrne, Chavis, Furnari, Girdhar, Hamburger, Jiang, Liu, Liu, et~al.]{grauman2022ego4d}
Kristen Grauman, Andrew Westbury, Eugene Byrne, Zachary Chavis, Antonino Furnari, Rohit Girdhar, Jackson Hamburger, Hao Jiang, Miao Liu, Xingyu Liu, et~al.
\newblock Ego4d: Around the world in 3,000 hours of egocentric video.
\newblock In \emph{Proceedings of the IEEE/CVF Conference on Computer Vision and Pattern Recognition}, pages 18995--19012, 2022.

\bibitem[Grauman et~al.(2024)Grauman, Westbury, Torresani, Kitani, Malik, Afouras, Ashutosh, Baiyya, Bansal, Boote, et~al.]{grauman2024ego}
Kristen Grauman, Andrew Westbury, Lorenzo Torresani, Kris Kitani, Jitendra Malik, Triantafyllos Afouras, Kumar Ashutosh, Vijay Baiyya, Siddhant Bansal, Bikram Boote, et~al.
\newblock Ego-exo4d: Understanding skilled human activity from first-and third-person perspectives.
\newblock In \emph{Proceedings of the IEEE/CVF Conference on Computer Vision and Pattern Recognition}, pages 19383--19400, 2024.

\bibitem[Graziano(2006)]{graziano2006organization}
Michael Graziano.
\newblock The organization of behavioral repertoire in motor cortex.
\newblock \emph{Annu. Rev. Neurosci.}, 29\penalty0 (1):\penalty0 105--134, 2006.

\bibitem[Guo et~al.(2023)Guo, Du, Shen, Lepetit, Alameda-Pineda, and Moreno-Noguer]{guo2023back}
Wen Guo, Yuming Du, Xi Shen, Vincent Lepetit, Xavier Alameda-Pineda, and Francesc Moreno-Noguer.
\newblock Back to mlp: A simple baseline for human motion prediction.
\newblock In \emph{Proceedings of the IEEE/CVF winter conference on applications of computer vision}, pages 4809--4819, 2023.

\bibitem[Hayhoe and Ballard(2005)]{hayhoe2005eye}
Mary Hayhoe and Dana Ballard.
\newblock Eye movements in natural behavior.
\newblock \emph{Trends in cognitive sciences}, 9\penalty0 (4):\penalty0 188--194, 2005.

\bibitem[Hernandez et~al.(2019)Hernandez, Gall, and Moreno-Noguer]{hernandez2019human}
Alejandro Hernandez, Jurgen Gall, and Francesc Moreno-Noguer.
\newblock Human motion prediction via spatio-temporal inpainting.
\newblock In \emph{Proceedings of the IEEE/CVF International Conference on Computer Vision}, pages 7134--7143, 2019.

\bibitem[Ho et~al.(2020)Ho, Jain, and Abbeel]{ho2020denoising}
Jonathan Ho, Ajay Jain, and Pieter Abbeel.
\newblock Denoising diffusion probabilistic models.
\newblock \emph{Advances in neural information processing systems}, 33:\penalty0 6840--6851, 2020.

\bibitem[Hu et~al.(2024)Hu, Schmitt, H{\"a}ufle, and Bulling]{hu2024gazemotion}
Zhiming Hu, Syn Schmitt, Daniel H{\"a}ufle, and Andreas Bulling.
\newblock Gazemotion: Gaze-guided human motion forecasting.
\newblock In \emph{2024 IEEE/RSJ International Conference on Intelligent Robots and Systems (IROS)}, pages 13017--13022. IEEE, 2024.

\bibitem[Jia et~al.(2022)Jia, Liu, and Rehg]{jia2022generative}
Wenqi Jia, Miao Liu, and James~M Rehg.
\newblock Generative adversarial network for future hand segmentation from egocentric video.
\newblock In \emph{European Conference on Computer Vision}, pages 639--656. Springer, 2022.

\bibitem[Jiang et~al.(2024)Jiang, Xie, Lin, Xu, Wan, Mandlekar, Fan, and Zhu]{jiang2024dexmimicgen}
Zhenyu Jiang, Yuqi Xie, Kevin Lin, Zhenjia Xu, Weikang Wan, Ajay Mandlekar, Linxi Fan, and Yuke Zhu.
\newblock Dexmimicgen: Automated data generation for bimanual dexterous manipulation via imitation learning.
\newblock \emph{arXiv preprint arXiv:2410.24185}, 2024.

\bibitem[Johansson and Flanagan(2009)]{johansson2009coding}
Roland~S Johansson and J~Randall Flanagan.
\newblock Coding and use of tactile signals from the fingertips in object manipulation tasks.
\newblock \emph{Nature Reviews Neuroscience}, 10\penalty0 (5):\penalty0 345--359, 2009.

\bibitem[Kerr et~al.(2025)Kerr, Hari, Weber, Kim, Yi, bonnen, Goldberg, and Kanazawa]{kerr2025eye}
Justin Kerr, Kush Hari, Ethan Weber, Chung~Min Kim, Brent Yi, tyler bonnen, Ken Goldberg, and Angjoo Kanazawa.
\newblock Eye, robot: Learning to look to act with a bc-rl perception-action loop.
\newblock In \emph{Proceedings of The 9th Conference on Robot Learning}, pages 3647--3664. PMLR, 2025.

\bibitem[Kim et~al.(2024)Kim, Pertsch, Karamcheti, Xiao, Balakrishna, Nair, Rafailov, Foster, Lam, Sanketi, et~al.]{kim2024openvla}
Moo~Jin Kim, Karl Pertsch, Siddharth Karamcheti, Ted Xiao, Ashwin Balakrishna, Suraj Nair, Rafael Rafailov, Ethan Foster, Grace Lam, Pannag Sanketi, et~al.
\newblock Openvla: An open-source vision-language-action model.
\newblock \emph{arXiv preprint arXiv:2406.09246}, 2024.

\bibitem[Konrad et~al.(2024)Konrad, Padmanaban, Buckmaster, Boyle, and Wetzstein]{Konrad2024}
Robert Konrad, Nitish Padmanaban, J.~Gabriel Buckmaster, Kevin~C. Boyle, and Gordon Wetzstein.
\newblock Gazegpt: Augmenting human capabilities using gaze-contingent contextual ai for smart eyewear.
\newblock \emph{arXiv preprint arXiv:2401.17217}, 2024.

\bibitem[Kratzer et~al.(2020)Kratzer, Bihlmaier, Midlagajni, Prakash, Toussaint, and Mainprice]{kratzer2020mogaze}
Philipp Kratzer, Simon Bihlmaier, Niteesh~Balachandra Midlagajni, Rohit Prakash, Marc Toussaint, and Jim Mainprice.
\newblock Mogaze: A dataset of full-body motions that includes workspace geometry and eye-gaze.
\newblock \emph{IEEE Robotics and Automation Letters}, 6\penalty0 (2):\penalty0 367--373, 2020.

\bibitem[Lai et~al.(2022)Lai, Liu, Ryan, and Rehg]{lai2022eye}
Bolin Lai, Miao Liu, Fiona Ryan, and James Rehg.
\newblock In the eye of transformer: Global-local correlation for egocentric gaze estimation.
\newblock \emph{British Machine Vision Conference}, 2022.

\bibitem[Lai et~al.(2024)Lai, Ryan, Jia, Liu, and Rehg]{lai2024listen}
Bolin Lai, Fiona Ryan, Wenqi Jia, Miao Liu, and James~M Rehg.
\newblock Listen to look into the future: Audio-visual egocentric gaze anticipation.
\newblock In \emph{European Conference on Computer Vision}, pages 192--210. Springer, 2024.

\bibitem[Land and Tatler(2009)]{land2009looking}
Michael Land and Benjamin Tatler.
\newblock \emph{Looking and acting: vision and eye movements in natural behaviour}.
\newblock Oxford University Press, 2009.

\bibitem[Land and Hayhoe(2001)]{land2001ways}
Michael~F Land and Mary Hayhoe.
\newblock In what ways do eye movements contribute to everyday activities?
\newblock \emph{Vision research}, 41\penalty0 (25-26):\penalty0 3559--3565, 2001.

\bibitem[Lappi and Mole(2018)]{lappi2018visuomotor}
Otto Lappi and Callum Mole.
\newblock Visuomotor control, eye movements, and steering: A unified approach for incorporating feedback, feedforward, and internal models.
\newblock \emph{Psychological bulletin}, 144\penalty0 (10):\penalty0 981, 2018.

\bibitem[Li et~al.(2024)Li, Zhu, Xie, Jiang, Seo, Pavlakos, and Zhu]{li2024okami}
Jinhan Li, Yifeng Zhu, Yuqi Xie, Zhenyu Jiang, Mingyo Seo, Georgios Pavlakos, and Yuke Zhu.
\newblock Okami: Teaching humanoid robots manipulation skills through single video imitation.
\newblock In \emph{8th Annual Conference on Robot Learning}, 2024.

\bibitem[Liu et~al.(2022)Liu, Tripathi, Majumdar, and Wang]{liu2022joint}
Shaowei Liu, Subarna Tripathi, Somdeb Majumdar, and Xiaolong Wang.
\newblock Joint hand motion and interaction hotspots prediction from egocentric videos.
\newblock In \emph{Proceedings of the IEEE/CVF Conference on Computer Vision and Pattern Recognition}, pages 3282--3292, 2022.

\bibitem[Lynch et~al.(2019)Lynch, Khansari, Xiao, Kumar, Tompson, Levine, and Sermanet]{lynch2019learning}
Corey Lynch, Mohi Khansari, Ted Xiao, Vikash Kumar, Jonathan Tompson, Sergey Levine, and Pierre Sermanet.
\newblock Learning latent plans from play.
\newblock In \emph{Conference on Robot Learning}, 2019.

\bibitem[Ma et~al.(2024)Ma, Ye, Hong, Guzov, Jiang, Postyeni, Pesqueira, Gamino, Baiyya, Kim, et~al.]{ma2024nymeria}
Lingni Ma, Yuting Ye, Fangzhou Hong, Vladimir Guzov, Yifeng Jiang, Rowan Postyeni, Luis Pesqueira, Alexander Gamino, Vijay Baiyya, Hyo~Jin Kim, et~al.
\newblock Nymeria: A massive collection of multimodal egocentric daily motion in the wild.
\newblock \emph{arXiv preprint arXiv:2406.09905}, 2024.

\bibitem[Mandlekar et~al.(2021)Mandlekar, Xu, Wong, Nasiriany, Wang, Kulkarni, Fei-Fei, Savarese, Zhu, and Mart\'{i}n-Mart\'{i}n]{robomimic2021}
Ajay Mandlekar, Danfei Xu, Josiah Wong, Soroush Nasiriany, Chen Wang, Rohun Kulkarni, Li Fei-Fei, Silvio Savarese, Yuke Zhu, and Roberto Mart\'{i}n-Mart\'{i}n.
\newblock What matters in learning from offline human demonstrations for robot manipulation.
\newblock In \emph{Conference on Robot Learning (CoRL)}, 2021.

\bibitem[Mao et~al.(2019)Mao, Liu, Salzmann, and Li]{mao2019learning}
Wei Mao, Miaomiao Liu, Mathieu Salzmann, and Hongdong Li.
\newblock Learning trajectory dependencies for human motion prediction.
\newblock In \emph{Proceedings of the IEEE/CVF international conference on computer vision}, pages 9489--9497, 2019.

\bibitem[Martinez et~al.(2017)Martinez, Black, and Romero]{martinez2017human}
Julieta Martinez, Michael~J Black, and Javier Romero.
\newblock On human motion prediction using recurrent neural networks.
\newblock In \emph{Proceedings of the IEEE conference on computer vision and pattern recognition}, pages 2891--2900, 2017.

\bibitem[Nijhawan and Wu(2009)]{nijhawan2009compensating}
Romi Nijhawan and Si Wu.
\newblock Compensating time delays with neural predictions: are predictions sensory or motor?
\newblock \emph{Philosophical Transactions of the Royal Society A: Mathematical, Physical and Engineering Sciences}, 367\penalty0 (1891):\penalty0 1063--1078, 2009.

\bibitem[Pavllo et~al.(2020)Pavllo, Feichtenhofer, Auli, and Grangier]{pavllo2020modeling}
Dario Pavllo, Christoph Feichtenhofer, Michael Auli, and David Grangier.
\newblock Modeling human motion with quaternion-based neural networks.
\newblock \emph{International Journal of Computer Vision}, 128:\penalty0 855--872, 2020.

\bibitem[Ren et~al.(2025)Ren, Sundaresan, Sadigh, Choudhury, and Bohg]{ren2025motion}
Juntao Ren, Priya Sundaresan, Dorsa Sadigh, Sanjiban Choudhury, and Jeannette Bohg.
\newblock Motion tracks: A unified representation for human-robot transfer in few-shot imitation learning.
\newblock \emph{arXiv preprint arXiv:2501.06994}, 2025.

\bibitem[Saran et~al.(2018)]{Saran2018}
A. Saran et~al.
\newblock Human gaze following for human-robot interaction.
\newblock In \emph{IEEE/RSJ International Conference on Intelligent Robots and Systems (IROS ’18)}, page ??–??, 2018.

\bibitem[Shadmehr and Krakauer(2008)]{shadmehr2008computational}
Reza Shadmehr and John~W Krakauer.
\newblock A computational neuroanatomy for motor control.
\newblock \emph{Experimental brain research}, 185:\penalty0 359--381, 2008.

\bibitem[Shadmehr et~al.(2010)Shadmehr, Smith, and Krakauer]{shadmehr2010error}
Reza Shadmehr, Maurice~A Smith, and John~W Krakauer.
\newblock Error correction, sensory prediction, and adaptation in motor control.
\newblock \emph{Annual review of neuroscience}, 33\penalty0 (1):\penalty0 89--108, 2010.

\bibitem[Wang et~al.(2021)Wang, Wang, Mandlekar, Fei-Fei, Savarese, and Xu]{wang2021generalization}
Chen Wang, Rui Wang, Ajay Mandlekar, Li Fei-Fei, Silvio Savarese, and Danfei Xu.
\newblock Generalization through hand-eye coordination: An action space for learning spatially-invariant visuomotor control.
\newblock In \emph{{2021 IEEE/RSJ International Conference on Intelligent Robots and Systems (IROS)}}, pages 8913--8920. IEEE, 2021.

\bibitem[Wolpert and Flanagan(2001)]{wolpert2001motor}
Daniel~M Wolpert and J~Randall Flanagan.
\newblock Motor prediction.
\newblock \emph{Current biology}, 11\penalty0 (18):\penalty0 R729--R732, 2001.

\bibitem[Yan et~al.(2023)Yan, Hu, Schmitt, and Bulling]{yan2023gazemodiff}
Haodong Yan, Zhiming Hu, Syn Schmitt, and Andreas Bulling.
\newblock Gazemodiff: Gaze-guided diffusion model for stochastic human motion prediction.
\newblock \emph{arXiv preprint arXiv:2312.12090}, 2023.

\bibitem[Zago et~al.(2009)Zago, McIntyre, Senot, and Lacquaniti]{zago2009visuo}
Myrka Zago, Joseph McIntyre, Patrice Senot, and Francesco Lacquaniti.
\newblock Visuo-motor coordination and internal models for object interception.
\newblock \emph{Experimental Brain Research}, 192:\penalty0 571--604, 2009.

\bibitem[Zeng et~al.(2021)Zeng, Florence, Tompson, Welker, Chien, Attarian, Armstrong, Krasin, Duong, Sindhwani, et~al.]{zeng2020transporter}
Andy Zeng, Pete Florence, Jonathan Tompson, Stefan Welker, Jonathan Chien, Maria Attarian, Travis Armstrong, Ivan Krasin, Dan Duong, Vikas Sindhwani, et~al.
\newblock Transporter networks: Rearranging the visual world for robotic manipulation.
\newblock In \emph{Conference on Robot Learning}, pages 726--747. PMLR, 2021.

\bibitem[Zhang et~al.(2017)Zhang, Teck~Ma, Hwee~Lim, Zhao, and Feng]{zhang2017deep}
Mengmi Zhang, Keng Teck~Ma, Joo Hwee~Lim, Qi Zhao, and Jiashi Feng.
\newblock Deep future gaze: Gaze anticipation on egocentric videos using adversarial networks.
\newblock In \emph{Proceedings of the IEEE conference on computer vision and pattern recognition}, pages 4372--4381, 2017.

\bibitem[Zheng et~al.(2022)Zheng, Yang, Mo, Li, Yu, Liu, Liu, and Guibas]{zheng2022gimo}
Yang Zheng, Yanchao Yang, Kaichun Mo, Jiaman Li, Tao Yu, Yebin Liu, C~Karen Liu, and Leonidas~J Guibas.
\newblock Gimo: Gaze-informed human motion prediction in context.
\newblock In \emph{European Conference on Computer Vision}, pages 676--694. Springer, 2022.

\bibitem[Zhu et~al.(2023)Zhu, Jiang, Stone, and Zhu]{zhu2023learning}
Yifeng Zhu, Zhenyu Jiang, Peter Stone, and Yuke Zhu.
\newblock Learning generalizable manipulation policies with object-centric 3d representations.
\newblock \emph{arXiv preprint arXiv:2310.14386}, 2023.

\bibitem[Zhu et~al.(2024)Zhu, Lim, Stone, and Zhu]{zhu2024vision}
Yifeng Zhu, Arisrei Lim, Peter Stone, and Yuke Zhu.
\newblock Vision-based manipulation from single human video with open-world object graphs.
\newblock \emph{arXiv preprint arXiv:2405.20321}, 2024.

\end{thebibliography}
